\documentclass[conference]{IEEEtran}
\IEEEoverridecommandlockouts
\usepackage{cite}
\usepackage{amsmath,amssymb,amsfonts}
\usepackage{graphicx}
\usepackage{textcomp}
\usepackage{xcolor}
\def\BibTeX{{\rm B\kern-.05em{\sc i\kern-.025em b}\kern-.08em
    T\kern-.1667em\lower.7ex\hbox{E}\kern-.125emX}}
    
\usepackage{algorithm}
\usepackage{algpseudocode}
\usepackage{tablefootnote}
\usepackage{multirow}
\usepackage{adjustbox}

\begin{document}

\bibliographystyle{IEEEtran}

\title{Fast Training of Deep Neural Networks \\ Robust to Adversarial Perturbations \\
\thanks{Research was sponsored by the United States Air Force Research Laboratory and was accomplished under Cooperative Agreement Number FA8750-19-2-1000. The views and conclusions contained in this document are those of the authors and should not be interpreted as representing the official policies, either expressed or implied, of the United States Air Force or the U.S. Government. The U.S. Government is authorized to reproduce and distribute reprints for Government purposes notwithstanding any copyright notation herein.}
}

\author{\IEEEauthorblockN{Justin Goodwin\textsuperscript{\textsection}}
\IEEEauthorblockA{\textit{MIT Lincoln Laboratory} \\
Lexington, MA, USA \\
jgoodwin@ll.mit.edu}
\and
\IEEEauthorblockN{Olivia Brown\textsuperscript{\textsection}}
\IEEEauthorblockA{\textit{MIT Lincoln Laboratory} \\
Lexington, MA, USA \\
olivia.brown@ll.mit.edu}
\and
\IEEEauthorblockN{Victoria Helus}
\IEEEauthorblockA{\textit{MIT Lincoln Laboratory} \\
Lexington, MA, USA \\
victoria.helus@ll.mit.edu}
}

\maketitle
\begingroup\renewcommand\thefootnote{\textsection}
\footnotetext{Equal contribution}
\endgroup

\begin{abstract}
Deep neural networks are capable of training fast and generalizing well within many domains.  Despite their promising performance, deep networks have shown sensitivities to perturbations of their inputs (e.g., adversarial examples) and their learned feature representations are often difficult to interpret, raising concerns about their true capability and trustworthiness. Recent work in adversarial training, a form of robust optimization in which the model is optimized against adversarial examples, demonstrates the ability to improve performance sensitivities to perturbations and yield feature representations that are more interpretable. Adversarial training, however, comes with an increased computational cost over that of standard (i.e., nonrobust) training,  rendering it impractical for use in large-scale problems. Recent work suggests that a fast approximation to adversarial training shows promise for reducing training time and maintaining robustness in the presence of perturbations bounded by the infinity norm.  In this work, we demonstrate that this approach extends to the Euclidean norm and preserves the human-aligned feature representations that are common for robust models.  Additionally, we show that using a distributed training scheme can further reduce the time to train robust deep networks. Fast adversarial training is a promising approach that will provide increased security and explainability in machine learning applications for which robust optimization was previously thought to be impractical.
\end{abstract}


\section{Introduction}

Despite the significant promise of machine learning for a wide array of domains, many machine learning models--notably deep neural networks~\cite{krizhevsky2012imagenet, lecun1998gradient, szegedy2015going, simonyan2014very, he2015delving}--have proven to be vulnerable to inputs that have been perturbed in a small but deliberate way, commonly referred to as adversarial examples or attacks~\cite{biggio2013evasion,szegedy2013intriguing, goodfellow2014explaining}.  Additionally, interpreting deep networks remains an open challenge~\cite{gunning2017explainable}. Robustness and interpretability are important characteristics for machine learning models used in safety- and security-critical applications~\cite{qayyum2020secure, qayyum2019securing, isakov2019survey}. 

Many approaches have been proposed to address the vulnerability posed by adversarial examples such as modifying the training data~\cite{gu2014towards, shaham2018understanding}, altering the network architecture~\cite{cisse2017parseval}, or by defensive distillation~\cite{papernot2016distillation}. Yet these defenses are often defeated by subsequent adversarial attack methods~\cite{carlini2016defensive, carlini2017adversarial, carlini2017towards}. One defense that has emerged from this ``arms race" of attack and defense development is the use of adversarial training (first introduced in~\cite{goodfellow2014explaining}, with its connection to robust optimization discussed in~\cite{shaham2018understanding, madry2017towards, sinha2017certifiable}).  The most common form of adversarial training uses robust optimization with projected gradient descent (PGD) to generate adversarial examples within the training loop~\cite{madry2017towards}. 

In addition to improving robustness to adversarial attacks, when used with the Euclidean norm, adversarial training imposes a prior that is closely aligned with human visual perception, resulting in trained networks with more interpretable feature representations~\cite{ilyas2019adversarial,engstrom2019learning, engstrom2019a}. Adversarial training can therefore be used to not only increase the security of machine learning systems, but also help in scenarios where explanations are needed to increase operator trust.

Unfortunately, generating adversarial examples within the training loop comes with a significant computational cost. This increased cost may be hindering progress in robust model development and application of it to complex, real world problems. Recent work, however, has demonstrated that fast adversarial training may be possible without sacrificing robustness~\cite{shafahi2019adversarial,wong2020fast}.  Additionally, distributed computing has been shown to speed up training time significantly for many large scale, deep learning problems~\cite{goyal2017accurate, samsi2019distributed}.

In this paper, we extend the work from~\cite{wong2020fast} by applying their fast adversarial training approach to the Euclidean norm.  We then train robust models using this approach with distributed training using multiple GPUs, and assess their training time, robustness, and learned feature representations using the CIFAR-10~\cite{krizhevsky2009learning} and Restricted ImageNet~\cite{ilyas2019adversarial} datasets.  We establish that fast adversarial training preserves robustness and interpretability, demonstrating promise for large-scale problems that necessitate machine learning solutions that are robust, explainable, and efficient to train.

\section{Background}
\subsection{Adversarial Perturbations}

Given a machine learning model (e.g., deep neural network), $f_\theta$, parameterized by $\theta$, an input example, $x$, and true label, $y$, an adversarial perturbation, $\delta$, is found by solving the following optimization problem:
\begin{equation}
    \max\limits_{\delta\in \mathcal{S}} \mathcal{L}(f_\theta(x + \delta), y) \label{eq:pertopt}
\end{equation}
where $\mathcal{L}$ is the loss function (e.g., cross-entropy) and $\mathcal{S}$ is the space of allowable perturbations, often constrained by an $\ell_p$-norm.

Introduced in~\cite{goodfellow2014explaining}, one of the first methods proposed to approximate the solution to (\ref{eq:pertopt}) assumes $\mathcal{S}=\{\delta: || \delta ||_\infty \leq \epsilon\}$ and is known as the Fast Gradient Sign Method (FGSM). Adversarial perturbations are generated by FGSM as follows: 
\begin{equation}
    \delta^* = \epsilon \cdot \mathrm{sign}(\nabla_x\mathcal{L}(f_\theta(x), y))
\end{equation}
where $\nabla_x\mathcal{L}$ is the gradient of the loss function with respect to the input. 

A more accurate approximation to (\ref{eq:pertopt}) extends FGSM by taking $K$ steps of size $\alpha$ and projecting the perturbation onto $\mathcal{S}$ at each iteration:
\begin{align}
    \delta^{(k+1)} &= \Pi_{\mathcal{S}(\delta)}\left(\delta^{(k)} + \alpha \cdot \nabla_x\mathcal{L}(f_\theta(x+\delta^{(k)}), y)\right) \\
    \delta^* &= \delta^{(K)}
\end{align}
where $\Pi_{\mathcal{S}(\delta)}$ represents the projection of $\delta$ onto $\mathcal{S}$. This multi-step approach is known as projected gradient descent\footnote{More precisely, projected gradient descent on the negative loss function.} (PGD)~\cite{madry2017towards}. The initial perturbation for PGD, $\delta^{(0)}$, can be started from the origin, or sampled randomly from $\mathcal{S}$.  

PGD is able to produce a perturbation that is closer to the optimal solution of (\ref{eq:pertopt}) compared to single-step FGSM by taking multiple steps of gradient descent, and converges to the optimal solution as $K\rightarrow\infty$. While FGSM assumes the infinity norm, PGD is agnostic to the choice of $\mathcal{S}$, and is thus a more general formulation for an adversarial attack.

\subsection{Adversarial Training}

To increase the robustness of deep neural networks, adversarial training (first suggested in~\cite{goodfellow2014explaining} and formalized in~\cite{madry2017towards}) adds adversarial examples into the training procedure by re-formulating standard training into a robust min-max optimization problem: 
\begin{equation} \label{eq:minmax}
    \min\limits_\theta \sum_{i\in \mathcal{D}} \max\limits_{\delta\in\mathcal{S}}\mathcal{L}(f_\theta(x_i + \delta), y_i)
\end{equation}
where $\mathcal{D}$ is the training dataset. Similar to standard training, the outer minimization is solved at each iteration of training by updating the model weights, $\theta$, using an optimizer such as stochastic gradient descent (SGD), but with the loss computed on the adversarially-perturbed inputs:
\begin{equation} \label{eq:sgd}
    \theta \leftarrow \theta - \eta \cdot \frac{1}{N} \sum_{i=1}^N \nabla_\theta \mathcal{L} (f_\theta (x_i + \delta_i^*),y_i)
\end{equation}
where $N$ is the number of data samples used for the stochastic update (e.g., batch size), and $\eta$ is the learning rate. Adversarial training using FGSM was shown to be insufficient when tested against a stronger adversary such as PGD~\cite{madry2017towards}.  Thus, adversarial training with PGD is the preferred method for training robust models.

\subsection{Robust Feature Representations}

Approaches for increasing the interpretability of deep neural networks are often aimed at improving the learned feature representations~\cite{bengio2013representation, goodfellow2016deep}. In~\cite{ilyas2019adversarial}, the authors claim that adversarial perturbations are not bugs in the deep learning model, but nonrobust features that the model has found useful for maximizing accuracy during the standard training procedure.  They show that robust models trained with PGD and the Euclidean norm tend to learn more robust feature representations compared to standard models, and these robust features tend to align more closely with human cognition.  From a human perception standpoint, inputs that are close in input space should yield feature representations that are also close: 
\begin{equation}
    ||x - x'||_2 \leq \epsilon \implies ||f_\theta^R(x)-f_\theta^R(x')|| \leq C \cdot \epsilon
\end{equation}
where $f_\theta^R(x)$ is the feature representation of $x$ extracted by the model (e.g., the penultimate layer of a neural network) and $C$ is a constant.  This expression is very similar to the definition of adversarial robustness, so it follows that adversarial training provides a way to enforce a prior on the model for learning more human-aligned features~\cite{engstrom2019learning,ilyas2019adversarial}.

This insight opens up the doors to a number of remarkable observations and powerful tools that can be used on adversarially-trained models: inverting similar feature representations leads to perceptually similar inputs; visualizing the images that maximally-activate representation nodes reveal patterns that are recognizable by humans; performing large, targeted PGD attacks can be used to debug model errors or add features to an input that resemble specific classes~\cite{engstrom2019learning,santurkar2019image}.  We will use a sampling of these tools to qualitatively demonstrate that our robust models also have these properties.

\subsection{Fast Adversarial Training}

Using PGD during adversarial training is expensive due to the many gradient computations required to solve the inner maximization in (\ref{eq:minmax}).  Thus, scaling adversarial training to larger datasets and more complex problems is impractical without significant computational resources. 

Alternative approaches to multi-step PGD are being explored to speed up adversarial training. One such approach, referred to as ``free" adversarial training~\cite{shafahi2019adversarial}, proposes using a single backwards pass to simultaneously solve for the input perturbation and update the model parameters. While this prevents multiple perturbation steps for a given pass, the method overcomes this by training with the same mini-batch $T$ times so that each input is allowed multiple adversarial updates. Additionally, the optimal perturbations from one mini-batch are used as an initialization for the next mini-batch. The total number of epochs can be reduced by a factor of $T$ to make computational cost around the same as that of standard training.  Empirical results show that a robust model trained with free adversarial training performs just as well against PGD-generated attacks.

In~\cite{wong2020fast}, the authors hypothesize that the primary driver of the success of free adversarial training is the initialization of the perturbation from the previous mini-batch, which they claim serves as a form of randomization that allows a simpler attack to achieve similar robustness to PGD.  With this hypothesis,~\cite{wong2020fast} revisits FGSM for adversarial training, with the simple modification of randomly initializing the perturbation before taking the gradient step:
\begin{align}
    \delta^0 &\sim \mathrm{Uniform}(-\epsilon,\epsilon)  \\
    \delta^1 &= \delta^0 + \alpha \cdot \mathrm{sign}(\nabla_x \mathcal{L}(f_\theta (x+\delta^0),y) ) \\
    \delta^* &= \max ( \min (\delta^1,\epsilon), -\epsilon).
\end{align}
The authors observe that the random initialization allows their ``fast" adversarial training approach to yield comparable performance to PGD-trained models, even when tested against strong adversaries, suggesting that the FGSM approximation to the inner maximization of (\ref{eq:minmax}) may be sufficient for training robust models. Thus, there may be no need to sacrifice robustness for lower computational cost.

Both of the aforementioned fast training approaches assume a constraint on the infinity norm when training their robust models.  We will extend the approach from~\cite{wong2020fast} to the Euclidean norm, to explore if the conclusions hold when considering other norm-bounded constraints, and to investigate if human-aligned feature representations are attainable at reduced computational cost.

\section{Fast Adversarial Training with Constraint on Euclidean Norm}

Fast adversarial training with the Euclidean norm is completed as follows.  First, define the space of allowable perturbations as the $\ell_2$-ball, $\mathcal{S}_{2, \epsilon}=\{\delta: || \delta ||_2 \leq \epsilon\}$. An adversarial perturbation is solved for using a single step of PGD with random initialization sampled uniformly from $\mathcal{S}_{2, \epsilon}$: 
\begin{align}
    \delta^{(0)} &\sim \mathrm{Uniform}(\mathcal{S}_{2, \epsilon}) \label{eq:sample} \\
    \delta^{(1)} &= \delta^{(0)} + \alpha \cdot \frac{ \nabla_x \mathcal{L}(f_\theta (x+\delta^{(0)}),y) }{|| \nabla_x \mathcal{L} (f_\theta (x+\delta^{(0)}),y) ||_2} \label{eq:update} \\
    \delta^* &= \left\{ 
    \begin{array}{cl} 
        \epsilon \cdot \delta^{(1)} / || \delta^{(1)} ||_2 &\mbox{ if $|| \delta^{(1)} ||_2 > \epsilon$} \\
        \delta^{(1)} &\mbox{ otherwise}
    \end{array} \right. \label{eq:project}
\end{align}
where the gradient is normalized such that $\alpha$ controls the size of the step in (\ref{eq:update}).

The solution to the robust optimization problem in (\ref{eq:minmax}) is approximated by alternating between computing these fast perturbations using (\ref{eq:sample})--(\ref{eq:project}), and updating the model weights according to (\ref{eq:sgd}).  This process is described in Algorithm~\ref{alg:fast}, where training data is divided into $M$ mini-batches of size $N$ and training repeats for $E$ epochs. In practice, Lines 3--9 of Algorithm~\ref{alg:fast} are executed over the entire batch in parallel.

\begin{algorithm}
    \caption{Fast adversarial training with perturbations constrained by the $\ell_2$-norm.  Inputs are assumed to be mapped to the $[0,1]$ domain.} \label{alg:fast}
    \begin{algorithmic}[1]
        \For {$e = 1,\ldots, E$}
            \For {$m = 1,\ldots,M$}
                \For {$i = 1,\ldots,N$}
                
                    \State Solve for adversarial perturbation:
                    
                    \State $\delta_i \sim \mathrm{Uniform}(\mathcal{S}_{2, \epsilon})$
                    
                    \State $\delta_i = \delta_i + \alpha \cdot \frac{ \nabla_x \mathcal{L}(f_\theta (x_i + \delta_i),y) }{|| \nabla_x \mathcal{L} (f_\theta (x_i + \delta_i),y_i) ||_2}$
                    
                    \State $\delta_i = \delta_i.\mathrm{renorm}(p=2,\mathrm{maxnorm}=\epsilon)$
                    
                    \State $\delta_i^* = \mathrm{clamp}(\delta_i,0-x_i,1-x_i)$
                \EndFor
                
                \State Update model weights:
                
                \State $\theta \leftarrow \theta - \eta \cdot \frac{1}{N} \sum_{i=1}^N \nabla_\theta \mathcal{L} (f_\theta (x_i + \delta_i^*),y_i)$
            \EndFor
        \EndFor
    \end{algorithmic}
\end{algorithm}

\section{Experiments}
We empirically compare fast adversarial training (i.e., 1-step PGD) to training with 7 steps of PGD for perturbations constrained by the Euclidean norm. Our models are trained on the MIT Supercloud\footnote{https://supercloud.mit.edu/}, which provides compute nodes composed of two NVIDIA Volta V100 Graphics Processing Units (GPUs), 20 Intel Xeon Gold 20-core Central Processing Units (CPUs), and 384GB of RAM. We compare training using a single compute node (total of 2 GPUs) to training using 4 compute nodes (8 GPUs) utilizing the distributed data parallel method implemented in PyTorch~\cite{NEURIPS2019_9015} along with PyTorch Lightning~\cite{falcon2019pytorch}.

We use two datasets for our experiments: CIFAR-10~\cite{krizhevsky2009learning} and Restricted ImageNet~\cite{ilyas2019adversarial}.  CIFAR-10 is composed of images of size 32x32 grouped into 10 classes (airplane, automobile, bird, cat, deer, dog, frog, horse, ship, and truck). The CIFAR-10 dataset contains 50,000 samples for training and 10,000 for validation. Restricted ImageNet, originally introduced in~\cite{ilyas2019adversarial}, groups subsets of classes from ImageNet \cite{russakovsky2015imagenet} into 10 super-classes (dog, cat, frog, turtle, bird, primate, fish, crab, and insect).  With images of size 256x256 (cropped to 224x224 for input to the models), Restricted ImageNet represents a higher resolution alternative to CIFAR-10.  Restricted ImageNet contains 257,735 samples for training and 10,150 for validation. 

For both datasets, we use a 50-layer residual network architecture~\cite{he2016deep}, which contains roughly 23 million parameters. We use Madrylab's robustness package~\cite{robustness} to train and evaluate our models.  All models are trained for 150 epochs, with global batch sizes of 256.  Performance on the validation set is evaluated every 5 epochs. We use SGD to optimize the model weights with an initial learning rate of 0.1, momentum of 0.9, and weight decays of 5e-4 and 1e-4 for CIFAR-10 and Restricted ImageNet, respectively.  Learning rates are reduced by 10x every 50 epochs. 

We use an adversarial step size of $\alpha=1.5\cdot\epsilon$ for generating fast adversarial perturbations with 1-step PGD (informed by recommendations from~\cite{wong2020fast}) and $\alpha=2.5\cdot\epsilon/K$ when using $K$-step PGD (the default in~\cite{robustness}), and perturbations are randomly initialized for both 1- and $K$-step PGD. CIFAR-10  and Restricted ImageNet models are trained to be robust to perturbations of size $\epsilon=1.0$ and $\epsilon=3.0$, respectively.

\subsection{Training Time}

Refer to Table~\ref{tab:times} for the training times for adversarial training with 7- and 1-step PGD compared to standard (i.e., no PGD) training for CIFAR-10 and Restricted ImageNet with 2 and 8 GPUs. Training times are reported for both the last epoch ($150$) and the ``best" epoch, which is the epoch that achieves the highest adversarial accuracy\footnote{Adversarial accuracy is the average accuracy on adversarially-perturbed samples from the test set} on the validation set. 

With roughly a 3x speedup on both datasets for 2-GPU, fast adversarial training with 1-step PGD significantly reduces the time to train an $\ell_2$-robust model compared to 7-step PGD, and is roughly only 2x slower than standard training. Requiring 7x the gradient computations per batch compared to standard training, 7-step PGD is roughly 7x longer than standard. Additional speedups in training time are evident when comparing 2- and 8-GPU training. For the 8-GPU case, fast adversarial training is approximately as fast as standard training, and training with 7-step PGD requires roughly twice as long as standard. 

\begin{table}
{\caption{Training Times in Hours (Relative to 2-GPU, No PGD)}\label{tab:times}}
\begin{adjustbox}{width=\columnwidth,center}
    \begin{tabular}{llllll}
    Dataset & Approach & \multicolumn{2}{c}{Last Epoch} &  \multicolumn{2}{c}{Best Epoch} \\
    &\  & 2-GPU & 8-GPU & 2-GPU & 8-GPU  \\ \hline
    \multirow{3}{1cm}{CIFAR} 
    & 7-Step PGD & 9.1 (7.4) & 3.0 (2.4) & 6.4 (5.2) & 1.3 (1.1) \\ 
    & 1-Step PGD & 2.9 (2.3) & 1.1 (0.9) & 1.4 (1.2) & 0.5 (0.5) \\
    & No PGD   & 1.2 (1.0) & 0.7 (0.6) & 0.9 (0.7) & 0.5 (0.4) \\ \hline 
    
    \multirow{3}{.1cm}{Restricted \newline ImageNet}
    & 7-Step PGD\tablefootnote{Training did not complete for 7-step PGD on Restricted ImageNet with 2 GPUs, so its time for the last epoch is an estimate and the time for the best epoch is unavailable.} & 127.5 (7.1) & 38.8 (2.2) & -- & 2.6 (0.2) \\
    & 1-Step PGD & 39.8 (2.2)  & 12.9 (0.7) & 17.3 (0.9) & 10.9 (0.6) \\ 
    & No PGD & 17.7 (1.0)    & 8.4 (0.5)  & 12.3 (0.7) & 5.5 (0.3) \\ \hline 
    
    \end{tabular}
\end{adjustbox}

\end{table} 

\begin{figure*}
\centerline{\includegraphics[height=2in]{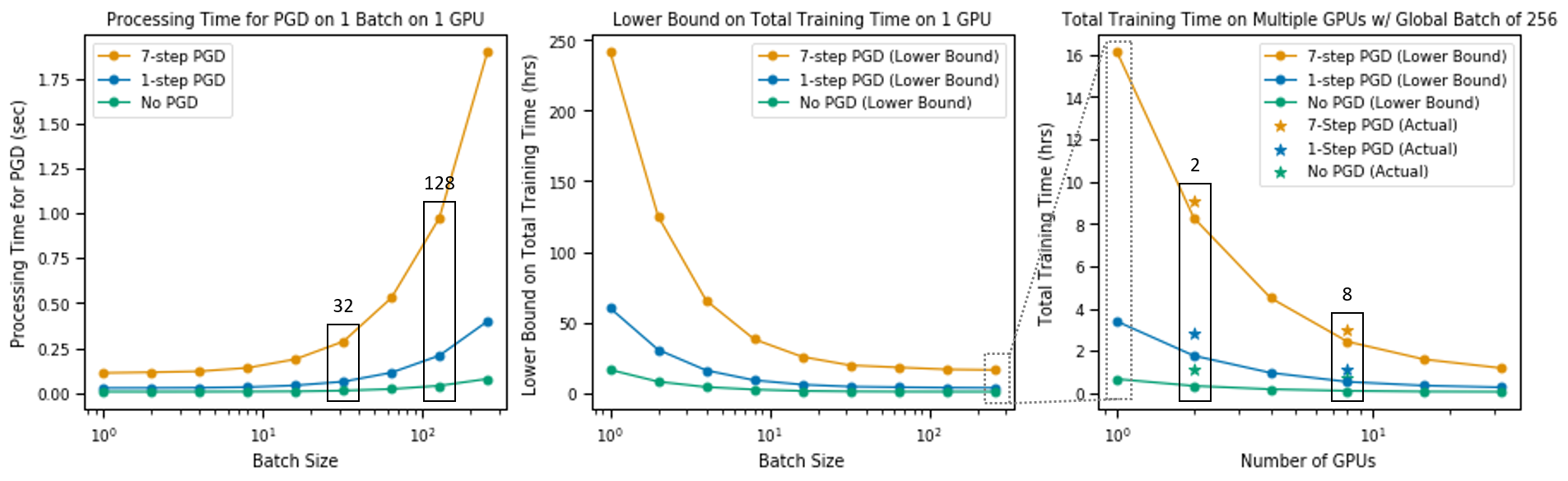}}
\caption{Analysis of execution time required for $\ell_2$-constrained PGD on CIFAR-10. Batch sizes and number of GPUs used to train the models in this paper are highlighted by the solid boxes. (Left) Average time for execution of a single instance of PGD using a single GPU and varying batch size. ``No PGD" is represented by executing a single forward pass. (Middle) Estimate of lower bound on total training time using a single GPU and varying batch size. (Right) Estimate of total training time given multiple GPUs and global batch size of 256. Actual training times from our experiments are indicated with stars.}  \label{fig:batch_times}
\end{figure*}

To better understand the drivers of the speedup when switching from 7- to 1-step PGD, we compute the time to execute PGD over varying batch sizes on a single GPU for the CIFAR-10 dataset. We also compute the time to execute a single forward pass through the network to serve as a proxy for standard (i.e., no PGD) training. These results are shown in the left plot of Figure~\ref{fig:batch_times}. Note that decreasing the batch size has diminishing returns on the PGD processing time.  

Next, we use these values to estimate the total time required for PGD executions over the entire training process (including $150$ epochs of training and $150/5=30$ epochs of validation), which represents a lower bound on the total training time. These estimates, shown in the middle plot of Figure~\ref{fig:batch_times}, demonstrate that increasing batch size reduces the overall training time. While smaller batches reduce execution time for a single instance of PGD, larger batches reduce the total number of iterations per epoch (and subsequently the total number of executions of PGD).  

There are diminishing returns with increasing batch size when computing PGD with a single GPU, thus a greater reduction in training time can be achieved by using distributed processing on multiple GPUs. This concept is illustrated in the right plot of Figure~\ref{fig:batch_times}, where we estimate a lower bound on the total training time for varying numbers of GPUs given a global batch size of 256. The training times from Table~\ref{tab:times} are also shown on the plot, and align nicely with our estimates.

Our analysis and empirical results on adversarial training time demonstrate that while reducing the number of steps for PGD does lead to significant reductions in training time, there is also a clear benefit to computing PGD on smaller batch sizes over multiple GPUs. Even for adversarial training with 7-step PGD, we see a large improvement in training time when moving from 2 to 8 GPUs. For this paper, we focus solely on the speedup due to the reduction in the number of gradient computations and changes in the batch size via distributed processing; however, we expect that other modifications, such as adjustments to the learning rate schedule and mixed-precision arithmetic (as demonstrated in~\cite{wong2020fast}), will enable even greater reductions in training time.

\subsection{Robustness}

While fast adversarial training with 1-step PGD clearly leads to improvements in training time, it is only useful for safety-critical applications if it achieves similar robustness to multi-step PGD. Thus, we assess the robustness of our trained models\footnote{For 7-step PGD for Restricted ImageNet, we use the pre-trained model from~\cite{engstrom2019learning}.} by computing their adversarial accuracy at varying levels of perturbation strength.  We evaluate the models at their best epoch, and use 20-step PGD with 10 random restarts for a range of $\epsilon$ values (assuming the $\ell_2$-norm) for this evaluation.

\begin{figure}
\centerline{\includegraphics[height=1.5in]{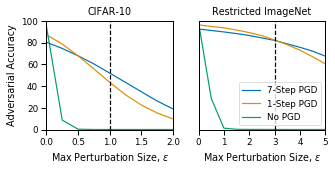}}
\caption{Adversarial accuracy as a function of maximum perturbation size ($\epsilon$) for CIFAR-10 and Restricted ImageNet models trained against $\ell_2$-norm bounded perturbations. The black, dotted lines indicate the $\epsilon$ used during adversarial training.}  \label{fig:perform_both}
\end{figure}

Refer to Figure~\ref{fig:perform_both} for these results.  When tested against smaller perturbations (lower $\epsilon$), the models trained with 1-step PGD achieve higher adversarial accuracy than those trained with 7-step PGD. At higher $\epsilon$, however, the 7-step PGD models achieve higher adversarial accuracy, and are thus considered to be more robust.  While 1-step PGD is not able to achieve an exact match to 7-step PGD, these results suggest that it is a useful approximation, as it achieves a much higher level of robustness compared to the standard (no PGD) model.

\subsection{Feature Representations}

As suggested in~\cite{engstrom2019learning}, adversarial robustness acts as a prior for learning human-aligned features, and we are interested in qualitatively assessing the feature representations learned via fast adversarial training to see if they preserve this notion of interpretability.  We study the feature representations of both standard and robust models using three visualization methods\footnote{The examples we show in this section are randomly sampled and are representative of the phenomena we consistently observe across many samples.}: direct feature visualizations, image interpolation, and large adversarial perturbations.

\subsubsection{Direct Feature Visualizations}

    The direct feature visualization for a given node, $i$, in the feature representation layer (e.g., penultimate layer of the network) is computed as follows: start from a seed image, $x_0$ (either random noise or a real image from the test set), and solve for an image $x^*$ that maximizes the activation at that node. Note that this can be solved for using PGD, where the loss term is now equal to the value of the activation at the node of interest, $f_\theta^{R,i}(x)$. In contrast to standard models, models trained for adversarial robustness tend to have features that exhibit clear, often recognizable patterns that persist across random initializations.
    
\begin{figure} 
\centerline{\includegraphics[height=5.5in]{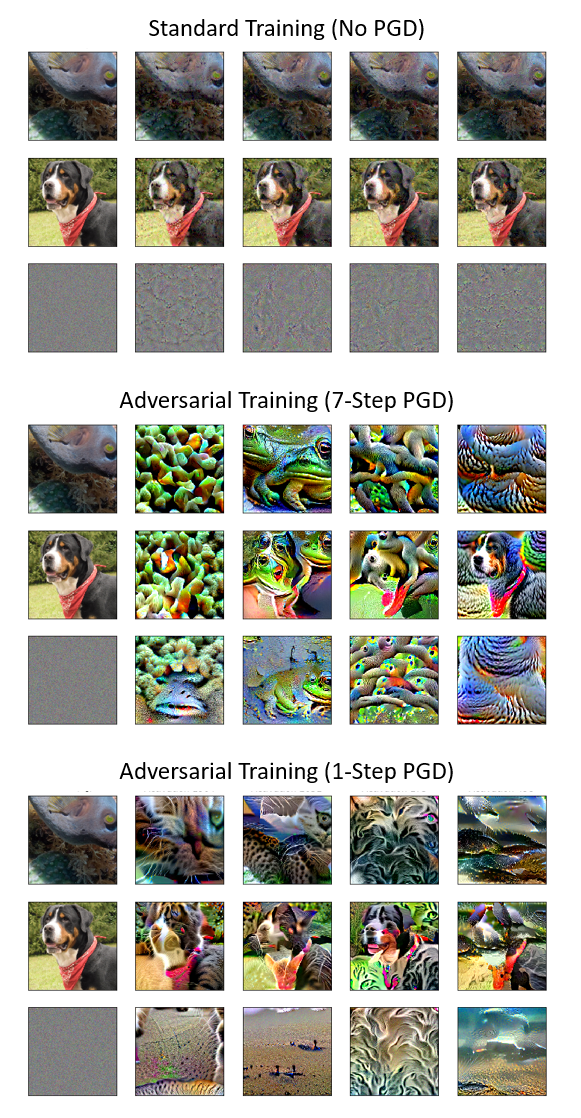}}
\caption{Direct feature visualizations for models trained with standard training and adversarial training with 1- and 7-step PGD.  The first column is the seed image, and the remaining 4 columns are the result of solving for the input that maximizes 5 randomly sampled nodes from the representation layer.} \label{fig:feat_viz}
\end{figure}

    Refer to Figure~\ref{fig:feat_viz} for direct feature visualizations for models trained on Restricted ImageNet with standard training and adversarial training using both 1- and 7-step PGD.  Similar to the features learned with 7-step PGD (albeit slightly less complex), the feature representations from 1-step PGD have distinct patterns that generally align with human perception, and both adversarial training approaches show clear improvement over standard training, whose features appear noise-like.
    
\subsubsection{Image Interpolation}
    Image interpolation is performed as follows: given two anchor images, $x_1$ and $x_2$, and an interpolation parameter, $\lambda$, solve for the $\lambda$-interpolated image, $x_{\lambda}$, by minimizing the distance between the current interpolation and the desired interpolation in the representation space: 
    \begin{equation}
        \min_{x_\lambda} \| (\lambda f_\theta^R(x_1) - (1-\lambda)f_\theta^R(x_2)) - f_\theta^R(x_\lambda)\|_2.
    \end{equation}
    This results in an image that has a similar representation as the linearly interpolated points in representation space. A robust model will provide meaningful and interpretable representations where a nonrobust model cannot.  
    
\begin{figure} 
\centerline{\includegraphics[height=3.0in]{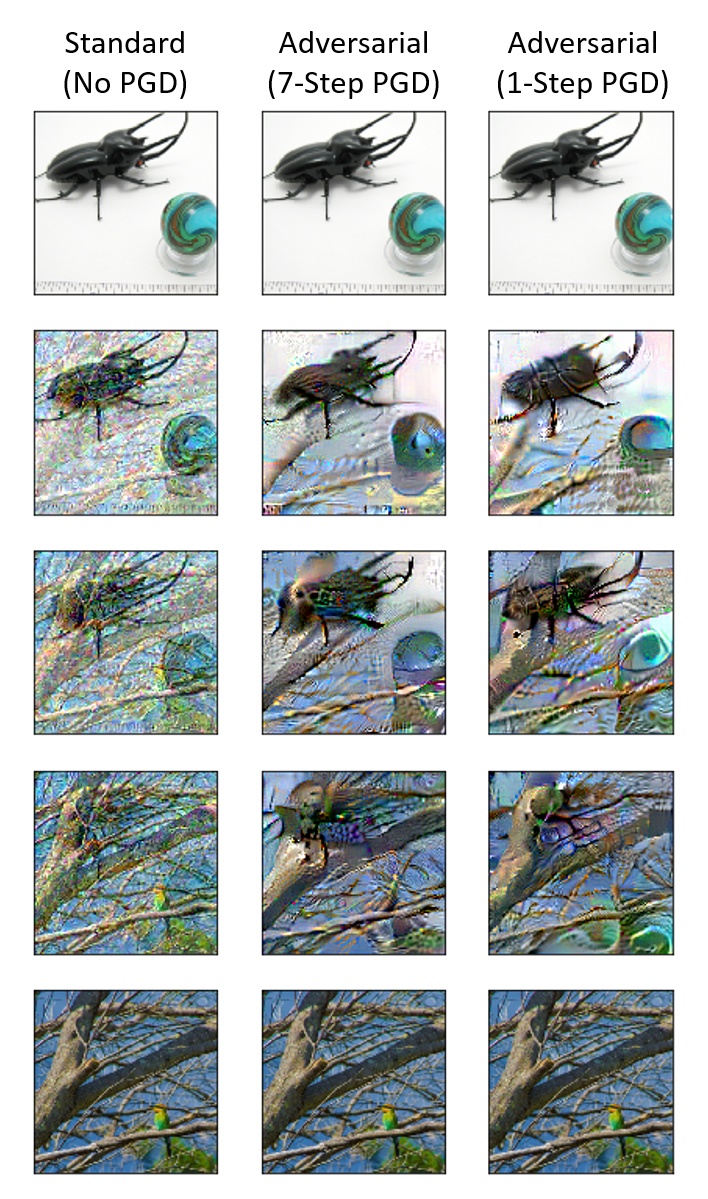}}
\caption{Image interpolation for models trained with standard training and adversarial training with 1- and 7-step PGD.  The first and last row are the anchor images, while the middle three rows are the interpolated images for varying levels of $\lambda$ for each of the three models.} \label{fig:img_interp}
\end{figure}

    In Figure~\ref{fig:img_interp}, we show the results of performing interpolation between two images for each of our models.  Similar to 7-step PGD, 1-step PGD has smoother transitions between the interpolations compared to the model trained with standard training.  This suggests that training with a fast, approximate solution to the inner objective in (\ref{eq:minmax}) still enables models to learn feature representations that exhibit an increased level of smoothness over those achieved with standard training.

\subsubsection{Large Adversarial Perturbations}
    Large adversarial perturbations are found by solving for a large perturbation, $\delta$, in the direction of a target class, $c$.  This is commonly accomplished using PGD with large values for $K$ and $\epsilon$ (e.g., $K=1000$ steps and $\epsilon=500$), where the objective is to minimize loss for the target class, $\mathcal{L}(f_\theta(x+\delta),c)$. While large adversarial perturbations for standard models tend to look like noisy versions of the original image, for adversarially-robust models, these perturbed images appear (to humans) similar to the target class.
    
\begin{figure} 
\centerline{\includegraphics[height=1.7in]{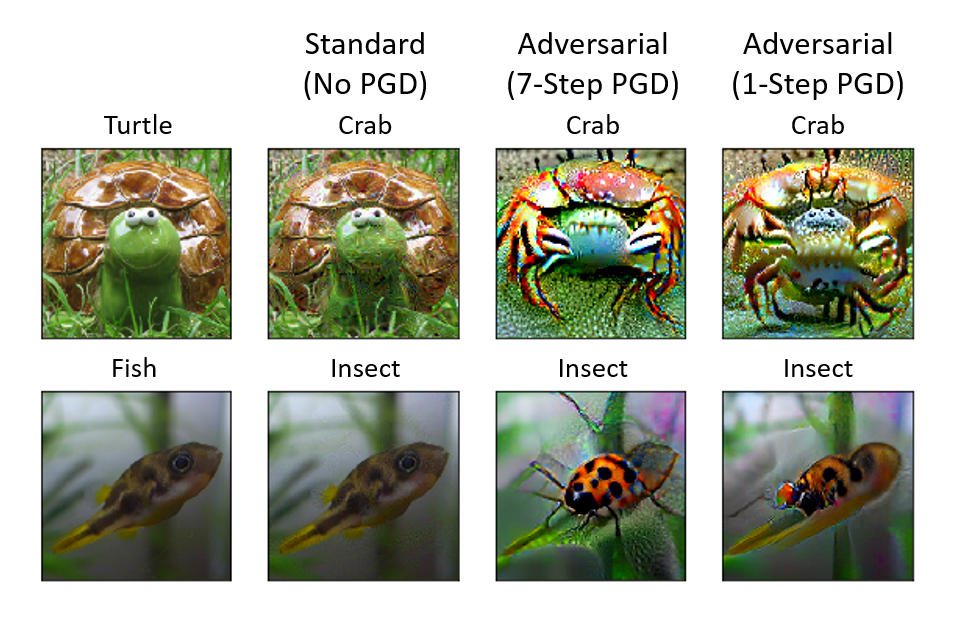}}
\caption{Large adversarial perturbations for models trained with standard training and adversarial training with 1- and 7-step PGD.  The first column is the original images with their true labels (turtle and fish), while the remaining columns show the adversarially-perturbed images that have been perturbed in the direction of the target class (crab and insect) for each model.} \label{fig:img_perturb}
\end{figure}

    Figure~\ref{fig:img_perturb} depicts images given large adversarial perturbations for our three models.  The perturbed images for the model trained with 1-step PGD visually resemble the target class, and show stark improvement over those for the model trained via standard training, whose perturbed images appear almost indistinguishable from the original class. The 1-step PGD images are, perhaps, slightly less distinct compared to the images for the model trained with 7-steps of PGD. 

\section{Discussion}
Our experiments demonstrate that fast adversarial training with random initialization and 1-step PGD significantly reduces the time to train models that are robust to adversarial perturbations constrained by the Euclidean norm, and is able to achieve robustness and feature representations that are similar to models trained using multi-step PGD. Future work will include extending this approach to additional distance measures (e.g., Wasserstein), and designing training schemes that use a combination of the two methods (e.g., 1-step PGD during earlier epochs, where an approximate solution may be sufficient, and multi-step PGD in later epochs).

We find that training time can be further reduced by utilizing distributed training, where multiple GPUs are used to process smaller batches of data.  We see a large reduction in training time from 2 to 8 GPUs for 7-step PGD, and future work will include designing distributed training regimes to improve the efficiently of multi-step PGD.  Such regimes will be needed for applications that require higher levels of robustness.

The techniques discussed in this paper can be useful tools for increasing the level of experimentation with robust models. For example, fast adversarial training and distributed processing may enable robust optimization to be run on large-scale problems for which multi-step PGD is currently prohibitive.  Additionally, fast adversarial training can be used for initial experiments when testing new techniques in a ``back-of-the-envelope" fashion, before using multi-step PGD for training the final model. Due to its additional ability to learn human-aligned features, fast adversarial training with the Euclidean norm may also be a great resource for the explainable AI community.

\section{Conclusion}
We extend the fast adversarial training approach from~\cite{wong2020fast} to the Euclidean norm, and find it to be a useful approximation to traditional robust optimization with multi-step PGD. By leveraging distributed training with multiple GPUs, we achieve further reductions in training time. Neural networks that are robust, interpretable, and quick to train will be important as deep learning is increasingly applied to large-scale, safety-critical problems.

\section*{Acknowledgment}

The authors would like to thank Rajmonda Caceres, Jeremy Kepner, Lori Layne, John Radovan, and Stephen Relyea for their feedback and support in conducting this research.

\bibliography{references}

\end{document}